# AdaptiveMDL-GenClust: A Robust Clustering Framework Integrating Normalized Mutual Information and Evolutionary Algorithms

**Hoda Jahani, Fabio Zamio**

Department of Computer Engineering, Azad University, IR

H.jahani@gmail.com

## Abstract

Clustering algorithms are pivotal in data analysis, enabling the organization of data into meaningful groups. However, individual clustering methods often exhibit inherent limitations and biases, preventing the development of a universal solution applicable to diverse datasets. To address these challenges, we introduce a robust clustering framework that integrates the Minimum Description Length (MDL) principle with a genetic optimization algorithm. The framework begins with an ensemble clustering approach to generate an initial clustering solution, which is then refined using MDL-guided evaluation functions and optimized through a genetic algorithm. This integration allows the method to adapt to the dataset's intrinsic properties, minimizing dependency on the initial clustering input and ensuring a data-driven, robust clustering process. We evaluated the proposed method on thirteen benchmark datasets using four established validation metrics: accuracy, normalized mutual information (NMI), Fisher score, and adjusted Rand index (ARI). Experimental results demonstrate that our approach consistently outperforms traditional clustering methods, yielding higher accuracy, improved stability, and reduced bias. The method's adaptability makes it effective across datasets with diverse characteristics, highlighting its potential as a versatile and reliable tool for complex clustering tasks. By combining the MDL principle with genetic optimization, this study offers a significant advancement in clustering methodology, addressing key limitations and delivering superior performance in varied applications.

**Keywords:** Normalized mutual information, Genetic optimization, unsupervised clustering.

## INTRODUCTION

Cluster analysis is a fundamental tool in modern data science, providing critical insights during the preprocessing stage of large-scale data analysis. This technique is particularly valuable when dealing with high-dimensional datasets, which can consist of tens, hundreds, or even thousands of attributes. By organizing data samples into clusters based on their similarities, clustering allows researchers to uncover hidden patterns and relationships in unlabeled data. The core principle of clustering is to group samples with similar attributes into the same cluster while ensuring that samples in different clusters exhibit distinct characteristics [1-4]. Over the years, clustering has found applications across a diverse range of fields, including bioinformatics, pattern recognition, machine learning, data mining, and image processing [3, 5-7].

Clustering plays a pivotal role in fields like bioinformatics, where gene expression data is often analyzed using clustering methods. By partitioning genes into clusters, researchers can infer relationships between genes, predict unknown functions, and better understand biological processes. For example, genes with similar functional roles or those participating in the same genetic pathways are typically grouped within the same cluster. This not only aids in exploring the genetic landscape but also facilitates advancements in precision medicine and disease diagnostics [8].

Beyond bioinformatics, clustering serves as a cornerstone for analyzing unlabeled data in diverse contexts. Its importance can be summarized in five key areas [9, 12]:

1. Facilitating Data Labeling: In many real-world scenarios, labeling datasets manually is prohibitively expensive and time-consuming. Clustering enables researchers to group similar data points, reducing the effort required for manual labeling.

2. Reverse Analysis: In data mining, clustering allows researchers to identify patterns and relationships in large, unlabeled datasets. Once clusters are identified, human observers can interpret and label the results for further analysis.

3. Adaptability to Change: In dynamic environments, such as seasonal food classification or adaptive recommendation systems, clustering can track evolving data attributes, ensuring that models remain relevant and accurate over time.

4. Feature Extraction: Clustering helps identify key attributes or features in datasets, enabling efficient data representation and improved performance in downstream machine learning tasks.

5. Structural Insight: By revealing the inherent structure and relationships within data, clustering provides valuable insights that inform decision-making and exploratory analysis.

With the explosion of big data across various domains, understanding and analyzing massive, high-dimensional datasets has become increasingly important. As a result, clustering has emerged as a crucial technique for making sense of complex data. However, the effectiveness of clustering largely depends on the choice of algorithm, which poses several challenges. Most existing clustering algorithms are broadly classified into hierarchical clustering and partitional clustering methods, each with its strengths and limitations [10].

Hierarchical clustering organizes data into a tree-like structure, or dendrogram, which represents the nested relationships among data points. Clusters are formed by cutting the dendrogram at a specific level. This approach can be further divided into two strategies: divisive and agglomerative clustering [11].

- Divisive Clustering: Begins with all data points in a single cluster and iteratively splits them into smaller clusters until each point forms its own cluster.

- Agglomerative Clustering: Starts with each data point as its own cluster and successively merges the closest clusters until all points are grouped into a single cluster.

While hierarchical clustering provides valuable insights into the data structure, it suffers from high computational complexity, making it unsuitable for large-scale datasets [11].

Partitional clustering methods, such as k-means, k-medoids, Forgy, and Isodata, divide the dataset into a predetermined number of clusters based on distance metrics or optimization criteria [13, 14]. Among these, k-means is widely used due to its simplicity, low computational cost, and ability to handle large datasets. However, k-means has several known limitations:

- Sensitivity to Initialization: Poor initialization of cluster centroids can lead to suboptimal results.

- Sensitivity to Outliers: Noise or outliers in the dataset can significantly skew clustering outcomes.

- Input Layout Dependency: The algorithm's performance depends heavily on the spatial distribution of data.

- Result Variability: Different initializations may produce inconsistent clustering results.

Given these limitations, traditional clustering methods often struggle to address the complexities of high-dimensional and dynamic datasets.

Emergence of Ensemble Clustering

To overcome the inherent weaknesses of individual clustering algorithms, ensemble clustering has emerged as a robust alternative in recent years. Ensemble clustering combines the outputs of multiple clustering algorithms to generate more accurate and stable clustering results [25, 28-32]. By leveraging the diversity of individual clustering solutions, ensemble methods reduce the biases and dependencies associated with single algorithms. Ensemble clustering typically involves two main stages:

1. Generation of Input Clusters: Produces a diverse set of clustering solutions using different algorithms or configurations [30, 37].
2. Consensus Combination: Aggregates the input clusters to form a unified clustering result, minimizing inconsistencies and maximizing accuracy [30, 37, 41-44].

Despite its advantages, ensemble clustering faces challenges in effectively combining diverse input clusters, particularly when some inputs are highly inaccurate or contradictory. Addressing this limitation requires innovative methods to mitigate the dependency on input clusters while incorporating additional data-driven insights.

This study introduces a novel clustering approach that integrates the Minimum Description Length (MDL) principle with a genetic optimization algorithm to overcome the biases and limitations of existing methods. The proposed method begins with an initial solution generated through ensemble clustering. Using evaluation functions based on MDL and genetic optimization, the solution is iteratively refined to achieve more accurate and stable clustering results. Unlike traditional ensemble methods that rely solely on external input clusters, the proposed approach leverages both external information and intrinsic data properties, reducing dependency on initial inputs and enhancing robustness.

The proposed methodology has been evaluated on multiple standard datasets using comprehensive validation metrics, demonstrating its ability to produce high-quality clusters suitable for a wide range of applications.

**THE PROPOSED METHOD**

In this study, we introduce a novel clustering method called Genetic MDL, which integrates the Minimum Description Length (MDL) principle with a genetic optimization algorithm to address the challenges of traditional clustering methods. The Genetic MDL framework aims to overcome the limitations of existing approaches by combining the strengths of MDL, which balances model complexity and data fidelity, with the adaptability and efficiency of genetic algorithms for optimization.

The Genetic MDL approach operates through three key optimization stages:

1. EPMDLGAO: This stage employs the MDL principle to evaluate and optimize partitioning solutions. It focuses on ensuring that the clustering result represents the data with minimal encoding cost, effectively balancing simplicity and accuracy.
2. ABMDLGAO: This stage applies MDL-based adjustments to the clustering model, further refining the cluster assignments to reduce redundancy and improve consistency across the dataset.
3. EPAFGAO: This stage incorporates an enhanced genetic algorithm to optimize cluster configurations, ensuring robust convergence to high-quality solutions.

The proposed framework leverages these three stages in sequence to iteratively refine the clustering process, ensuring that the final result is stable, accurate, and free from biases introduced by initial conditions or input clusters.

An outline of the Genetic MDL framework is provided in Figure 1, which illustrates the interplay between the MDL principle and the genetic optimization algorithm in each stage.

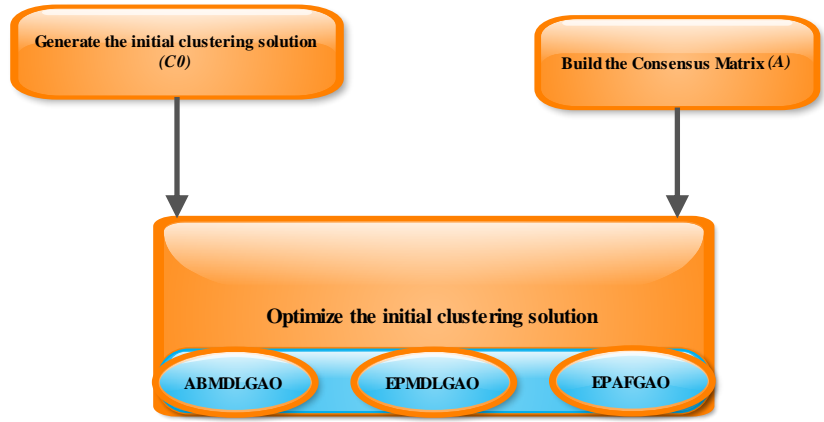

**Fig.1.** Framework of the proposed approach

The quasi-code of the proposed method is shown in Figure 2.

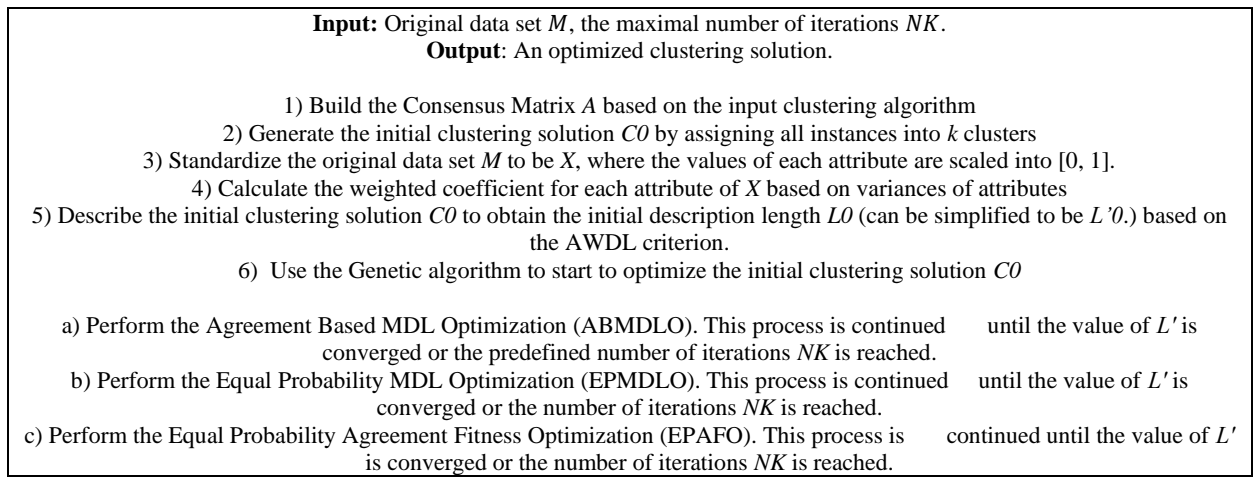

**Fig.2.** The quasi-code of the proposed method

## The Genetic MDL Framework

The Genetic MDL framework optimizes clustering by combining the Minimum Description Length (MDL) principle with genetic algorithms. It involves six key steps:

1. Formation of the Agreement Matrix (A): Constructs a matrix capturing consensus between clustering results from ensemble methods, forming the foundation for an initial solution.

2. Production of the Original Solution ($C_0$): Generates an initial clustering configuration by combining input clusters, ensuring diversity and robustness.

3. Normalization of Datasets: Scales data attributes to a uniform range, eliminating the influence of differing scales and ensuring consistent evaluation.

4. Calculation of Attribute Weight Coefficients: Assigns weights to attributes based on variance, prioritizing more informative attributes and reducing noise.

5. AWDL Description of the Original Solution: Evaluates the clustering solution using Attribute Weighted Description Length (AWDL) to balance model simplicity and data representation.

6. **Genetic Algorithm Multiple Optimization (GAMO):** Refines the clustering solution using genetic algorithm operations (selection, crossover, mutation) to achieve globally optimized, stable clusters.

**Formation of Agreement matrix (A)**

The Agreement matrix is made on the basis of the results of input clustering which appears in the form of a one-dimensional vector and each of them forms a matrix row. These individual clusters are selected by qualified people. For any data set, the length of output vector of each individual clustering is equal to the total number of items in the data set that shows the number of each sample's cluster. The Agreement Matrix is a square symmetric matrix with zero elements on the main diagonal. Each element of the Agreement matrix shows how many input clustering agreements agree with inserting two samples in a single cluster. Here the Agreement matrix is shown with variable $A$. $A$ is the square symmetric matrix $n \times n$, where $n$ is the number of samples in the data set.

Figure 3 shows the Agreement matrix formation algorithm.

```
for i=1:EnsembleSize
    IDX=kmeans(X,ClusterNum);
    out=[out,IDX];
end
for t=1:size(out,2)
    for i=1:size(out,2)
        B=zeros(size(out,1),1);
        for j=1:size(out,1)
            if out(j,t)==i
                B(j,1)=1;
            end
        end
        H=[H,B];
    end
end
A=H*H';
```

**Fig.3.** Agreement matrix formation algorithm

The formation of the Agreement Matrix (A) can be explained through an example. Consider three individual clustering algorithms: G1G_1G1, G2G_2G2, and G3G_3G3. Each algorithm clusters the dataset XXX, consisting of 10 samples, into three clusters. The clustering results are represented in vector form, where each element indicates the cluster assignment of a specific sample.

For instance, the clustering results are as follows:

Output G1 = {1 , 2, 1 , 1, 1, 3 , 3, 2 , 3, 1 }

Output G2 = {3 , 2, 3 , and 1, 2, 1 , 1, 3 , 2, 3 }

Output G3 = {2 , 1, 1, 3, 3 , 2, 1, 1, 2, 2 }

Results of G1, G2 and G3 clusters are displayed in the form of matrix Z. Then matrix Z is written in the form of matrix H.

The consensus matrix $A$ is obtained by equation 1.

(1) $$A = H \times H$$

In the above equation, $H'$ is the transpose of matrix $H$. In the above example, matrix $A$ is equal to:

**Production of the original solution (C0)**

There are various techniques to produce the original solution, each offering distinct advantages depending on the characteristics of the dataset. One common approach involves running the K-means clustering algorithm multiple times (e.g., five iterations) and combining its results with hierarchical clustering methods such as Linkage Average, Linkage Complete, Linkage Ward, and Linkage Single, each executed once. The total number of iterations is set to an odd number to facilitate consensus in the final solution. After aligning the cluster labels from these methods, the results are aggregated through an ensemble process to determine the initial clustering solution.

Another method involves random sampling from the dataset. For instance, a subset of the samples (e.g., 80%) is randomly selected, and the K-means algorithm is applied to this subset to create initial clusters. The remaining samples are then assigned to these clusters based on their similarity to cluster centroids, often measured using a distance metric such as Euclidean distance. This method is particularly efficient for large datasets, as it reduces computational complexity while maintaining accuracy. This approach has been employed in the present study to generate the original solution. The algorithm for generating the initial solution is illustrated in Figure 4.

Other techniques include density-based clustering, where methods like DBSCAN or OPTICS are used to identify dense regions in the data, which serve as initial clusters. Sparse points are treated as noise or assigned to the nearest cluster based on proximity. Additionally, model-based clustering, such as Gaussian Mixture Models (GMMs) or Expectation-Maximization (EM), can fit a probabilistic model to the data, providing cluster assignments as the initial solution. While effective for datasets with well-defined distributions, these methods may be computationally intensive for large datasets.

In this study, the random sampling with K-means clustering technique was selected due to its simplicity, efficiency, and ability to handle large datasets effectively, ensuring a robust starting point for subsequent optimization processes.

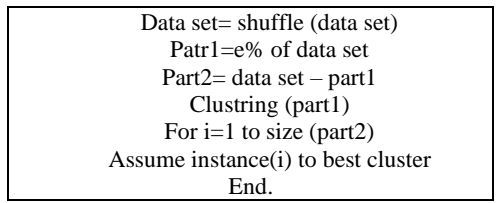

**Fig.4.** Initial solution generation algorithm

The following is an example to clarify this procedure.

Suppose that the data set *A* with 10 samples and the first part is equal to *A* '. After running the K-means algorithm on the data set *A* ', the clusters C1 and C2 are derived.

A = [1; 2; 3; 4; 5; 6; 7; 8]

A '= [1; 2; 3; 5; 6; 7]

C1 = {1; 2; 3}

C2 = {5; 6; 7}

The remaining samples (e.g., samples 4 and 8) are then assigned to clusters C1C_1C1 and C2C_2C2, respectively, based on their proximity to the centers of these clusters. Proximity is typically measured using a distance metric, such as the Euclidean distance, ensuring that each sample is allocated to the cluster with the closest centroid.

Another approach, referred to as the third method, is similar to the second method but replaces the K-means algorithm with hierarchical clustering algorithms such as Linkage Average, Linkage Complete, Linkage Ward, and Linkage Single. The process proceeds in the same way as the second method, where a subset of the dataset is clustered first, and the remaining samples are subsequently assigned to the generated clusters based on their similarity to cluster centers.

The fourth method combines the strengths of the second and third methods. It uses both the K-means algorithm and hierarchical clustering algorithms (Linkage methods) to cluster the selected subset of samples. This hybrid approach ensures that the clustering leverages the diverse perspectives of both partitional and hierarchical methods, offering a more robust initial solution. By integrating the outputs of these algorithms, the fourth method provides a more comprehensive representation of the data structure, particularly for datasets with complex distributions.

These methods collectively offer flexibility and adaptability for producing the original clustering solution, depending on the characteristics and requirements of the dataset. In this study, the second method, involving random sampling with the K-means algorithm, was chosen for its simplicity, efficiency, and scalability.

**Normalization of data sets**

Before applying AWDL, the data set M should be normalized and standardized to X. in normalization, the value of each attribute is set between 0 and 1. Each row of the data set M is an example and each column is an attribute. Equation 2 shows the normalization.

(2) $$X_{ij} = \frac{M_{ij} - \min(M_{ij})}{\max(M_{ij}) - \min(M_{ij})}, \quad 1 \leq i \leq n, 1 \leq j \leq n$$

In the above equation we have:

$X_{ij}$: represents the $j^{th}$ sample in the normalized data set X.

$M_{ij}$: represents the $j^{th}$ attribute of the $i^{th}$ sample in the main data set M.

n: is the total number of samples.

$\max(M_{ij})$ and $\min(M_{ij})$ are the highest and the lowest values in the $j^{th}$ attribute of the data set M.

**Calculation of the weight coefficient of the attributes**

Variance measurement is the degree of variation (difference) between the values of a variable. A higher variance represents more differences between the values of a variable. The higher the variance of a variable, the more weight it would have. Here we use the variance as the weight of each attribute.

To calculate the weight of each attribute, equations 3 to 5 are used.

(3) $$D_j = \frac{\sum_{i=1}^{n}(X_{ij} - \mu_j)^2}{n-1}$$

(4) $$\mu_j = \frac{\sum_{i=1}^{n} X_{ij}}{n}$$

(5) $$W_j = \frac{D_j}{\max(D_1, D_2, \ldots D_a)}$$

$D_j$: represents the variance of $j^{th}$ attribute for the samples of data set X.

n: is the number of samples in data set X.

$X_{ij}$: represents the value of $j^{th}$ attribute of the $i^{th}$ sample.

$\mu_j$: reflects the average value of $j^{th}$ attribute on n samples.

$w_j$: represents the weight the $j^{th}$ attribute.

a: is the total number of attributes.

**Description of the Initial Solution Based on the Attribute Weighted Description Length Criterion**

In this study, we employ the Attribute Weighted Description Length (AWDL) criterion, which builds upon the foundational principles of the Minimum Description Length (MDL) framework. MDL is a modern and widely applicable approach for comparative inference, offering a general solution to model selection problems. Its core principle states that:

Any pattern or regularity in the data can be used for compression.

In other words, if a dataset exhibits a certain structure or pattern, fewer bits or samples are required to describe the data than would be needed for a dataset with no discernible structure. The extent to which data can be compressed is directly proportional to the amount of order or regularity within the dataset. The key strength of the MDL approach is its versatility, as it can effectively describe various types of data.

Grunewald et al. (2004) demonstrated that MDL is a comprehensive approach for inference and model selection, outperforming other well-known methods, such as Bayesian statistical approaches, in terms of effectiveness and general applicability [40]. Today, MDL has been extended and applied in numerous research domains, including data clustering.

MDL in Clustering

The main concept behind using MDL in clustering is that the best clustering solution is the one that minimizes the description length of the dataset, rather than clustering solely based on similarity measures between samples. This approach ensures that clusters are formed in a way that optimally compresses the data, capturing its inherent structure.

To illustrate this concept, consider a dataset with NNN samples that need to be clustered into KKK clusters. The process involves two stages:

1. Formation of Initial Clusters: The KKK clusters are initialized, and the cluster centers are calculated based on the dataset.
2. Calculation of Distances: For each sample, the distance to the center of its assigned cluster is computed.

The quality of clustering is determined by the description length of the dataset. A clustering solution that minimizes the description length provides the best representation of the dataset. This methodology shifts the focus from traditional similarity-based clustering to a more holistic and information-theoretic perspective.

**Attribute Weighted Description Length Criterion**

The Attribute Weighted Description Length criterion is an extension of MDL, designed to incorporate attribute weighting into the clustering evaluation. Unlike standard MDL, which treats all attributes equally, AWDL assigns weights to attributes based on their variance, ensuring that more informative attributes contribute proportionally to the clustering process.

To calculate AWDL, the criterion LLL is applied, which consists of three components. The mathematical formulation of LLL is provided in Equation (6) and integrates the attribute weights, cluster structure, and overall dataset representation. The three components are:

1. Cluster Formation Cost: Captures the cost of forming clusters, including initializing cluster centers.
2. Cluster Assignment Cost: Represents the cost of assigning each sample to its respective cluster, considering its distance from the cluster center.
3. Compression Cost: Accounts for the compression achieved by clustering, balancing between simplicity and accuracy.

The AWDL criterion ensures that the clustering process is driven not only by the structural relationships within the data but also by the contribution of individual attributes, leading to a more robust and nuanced clustering solution.

(6)
$$L = \{Sm, Sd, W\}$$

In equation 6, W = [w1, w2, wa] is a vector that includes the weight of all attributes and since the vector W is always constant, AWDL can be easily expressed on the basis of criteria L ' and according to equation 7.

(7)
$$L' = \{Sm, Sd\}$$

Sm and Sd are described in the following.

Sm: represents the sum of the average value of all clusters of a partition and is defined based on Equation 8.

(8)
$$S_M = \sum_{p=1}^{k}(w_1 mp_1 + w_2 mp_2 + \cdots + w_a mp_a)$$

To describe Sm in equation 8 we have:

$a$: is the number of attributes (features) of the data set, and k is the number of clusters.

$mp_j$: represents the average value of $j^{th}$ attribute of the $p^{th}$ cluster.

$w_j$: is the weight $j^{th}$ attribute of the data set.

$sd$: is defined according to equation 9 and represents the weighted average deviation for all samples of each cluster in a partition. Here the standard deviation was used to measure the difference between the average and value of each sample in the cluster.

(9)
$$Sd = \sum_{p=1}^{k}\sum_{q=1}^{T}(w_1|X_{q1} - mp_1| + |X_{q2} - mp_2| + \cdots + |X_{qj} - mp_a|)$$

In the above equation we have:

a: is the number of attributes ( features ) of the data set, k is the number of clusters and T is the number of samples in each cluster.

$mp_j$: represents the average value of $j^{th}$ attribute of the $p^{th}$ cluster.

$X_{qj}$: is the value of j$^{th}$ attribute of the q$^{th}$ sample in p$^{th}$ cluster $(q = 1 \ldots T)$.

**Genetic algorithm Multiple Optimization framework (GAMO)**

GAMO framework consists of three separate optimization phases:

1. MDL optimizer based on Agreement Based MDL Genetic Algorithm Optimization (ABMDLGAO).
2. Equal Probability MDL Genetic Algorithm Optimization (EPMDLGAO).
3. Equal Probability Agreement Fitness Genetic Algorithm Optimization (EPAFGAO).

The performance each phase is investigated in the following.

**The optimization function ABMDLGAO**

Displacing the samples within clusters, this tries to find the best cluster to embed any sample so that the $L'$ criterion gets minimized (the $L'$ criterion was described above). The probabilities of sample selection for the displacement are not the same. To evaluate the probability of each sample's displacement, equations 10 to 13 are used. This process will continue until the value of $L'$ will converge or the iteration end condition will be reached.

Displacing the samples within clusters is a critical step in optimizing the clustering solution. This step aims to find the optimal cluster for each sample such that the $L'$ criterion—as described earlier—is minimized. By minimizing $L'$, the clustering process ensures that the overall description length of the dataset is reduced, leading to a more efficient and accurate representation of the data.

To achieve this, samples are iteratively evaluated for potential displacement to a different cluster. However, the probabilities for selecting samples for displacement are not uniform. Instead, they are determined based on their likelihood of improving the clustering solution, which is calculated using Equations (10) to (13). These equations incorporate factors such as the distance of a sample to the cluster centers and the contribution of its displacement to minimizing $L'$. This probability-based selection ensures that the optimization process focuses on the most impactful samples, reducing unnecessary computations and enhancing convergence speed.

The displacement process continues iteratively, with each iteration re-evaluating the clustering configuration. This iterative adjustment persists until one of two conditions is met:

1. Convergence of $L'$: The value of $L'$ stabilizes, indicating that further sample displacements will not significantly reduce the description length.
2. End Condition Reached: A predefined iteration limit or computational threshold is achieved, ensuring that the process terminates within practical runtime constraints.

By iteratively optimizing the placement of samples, this approach ensures a refined clustering solution that effectively balances accuracy, simplicity, and computational efficiency.

(10)
$$V1 = \frac{N_q \max(A_{iq})}{N_i \max(A_i)}$$

(11) $$V2 = \frac{\max(A_{iq})}{r}$$

(12) $$\begin{cases} 0 & \text{if} \max(A_{iq}) < \max(A_i) \\ \min(V1, V2) & \text{if} \max(A_{iq}) = \max(A_i) \end{cases}$$

(13) $$V_p = 1 - V$$

In the above equations we have:

$A$: the Consensus Matrix.

$V1$: integrated agreement rate

$A_{iq}$: agreed value of sample pairs that include sample i and are in the cluster q.

$max(A_{iq})$: maximum agreed value of $A_{iq}$

$N_q$: the number of maximum agreed value in $A_{iq}$.

$A_i$: the agreed value of all sample pairs that include sample $i$

$N_i$ : is the number of maximum values in $A_i$.

$max(A_i)$: is the maximum agreed value of $A_i$.

$V2$: is the simple agreed value

$r$: The number of input clusters

$V$: the sample's probability of not being selected to be displaced

$V_p$: the sample's probability of being selected to be displaced

**The Optimization function EPMDLGAO**

Similar to the ABMDLGAO function, this function aims to optimize sample placement within clusters to minimize the L'L'L' value. However, unlike ABMDLGAO, the probabilities for sample displacement are uniform.

**The Optimizer function EPAFGAO**

The objective function EPAFGAO serves as an agreed evaluation function, designed to identify solutions that maximize the value of the agreement objective function. The agreement function is mathematically defined using Equations (14) to (17).

(14) $$B = [\max(A) - \min(A)] \times 0.6 + \min(A)$$

(15) $$A' = A - B$$

$$(16) \quad f'(C_i) \begin{cases} \sum_{k=1}^{S_{i-1}} \sum_{q=k+1}^{S_i} (A'_{c_{ik},c_{iq}}) & \text{if } S_i > 1 \\ 0 & \text{otherwise} \end{cases}$$

$$(17) \quad F(C) = \sum_{i=1}^{m} f'(C_i)$$

In the above equations we have:

$A$: The Consensus Matrix

$B$ (Consensus Threshold): reward for the clusters whose agreement values are higher than B, and punishment for the clusters whose agreement values are less than B.

$max\ (A)$ and $min\ (A)$ are respectively the maximum and minimum values in matrix $A$.

$A'$: weighted agreement matrix obtained by the reduction of $B$ value from the consensus matrix $A$ and its main diagonal is zero.

$f'(C_i)$: the evaluation function of $i^{th}$ cluster of the total clusters of the initial solution.

$F\ (C)$: the evaluation function of the initial solution.

$S_i$: the number of samples in the $i^{th}$ cluster. If a cluster has only one sample (i.e. $S_i = 1$), the evaluation function of the cluster will be zero.

$C_{ik}$: the $k^{th}$ element of the $i^{th}$ cluster.

**EXPERIMENTAL RESULTS**

In this chapter, the experimental results of the evaluation of Genetic MDL across multiple datasets are presented. The datasets used in this study are sourced from the UCI Machine Learning Repository, a benchmark repository commonly utilized in clustering studies worldwide. The performance of Genetic MDL was assessed using various evaluation metrics, including Fisher's F-measure (F-measure), Normalized Mutual Information (NMI), Adjusted Rand Index (ARI), and Accuracy. Experimental results demonstrated that Genetic MDL exhibits satisfactory efficiency and robustness when applied to diverse datasets.

**Data Sets**

The Genetic MDL method was tested on 13 standard datasets to ensure a comprehensive evaluation. The datasets were selected to maximize diversity in terms of the number of classes, number of attributes, and number of samples, providing a robust basis for evaluating the method's generalizability and effectiveness. A summary of the datasets used is provided in Table 1, which highlights their key characteristics and ensures that the results reflect a wide range of data conditions.

**Table 1.** The used data sets and their features

| Data sets | # of classes | # of features | # of samples |
|---|---|---|---|
| balance_scale | 3 | 4 | 625 |
| Halfring | 2 | 2 | 400 |
| Iris | 3 | 4 | 150 |
| Nbalance_scale | 3 | 4 | 625 |
| Nbreast | 2 | 9 | 683 |
| Nbupa | 2 | 6 | 345 |
| Ngalaxy | 7 | 4 | 323 |
| Nglass | 6 | 9 | 214 |
| Nionosphere | 2 | 34 | 351 |
| NSAHeart | 2 | 9 | 462 |
| Nwine | 3 | 13 | 178 |
| NYeast | 10 | 4 | 1484 |
| Wine | 3 | 13 | 178 |

For all datasets, the number of clusters and the true labels of the samples were known beforehand. This allowed the percentage of correctly recognized samples to serve as the efficiency metric for the clustering approach. By addressing the correspondence between the obtained labels and the actual clusters, the error rate was calculated to evaluate the performance of the method.

**Validation measures**

To compare the efficiencies of different algorithms, four widely recognized evaluation criteria were employed: Accuracy, NMI, Fisher's F-measure (F-measure), and ARI. The definitions and significance of these criteria are explained below.

**Fisher measure**

The Fisher's F-measure is a criterion that combines accuracy and precision into a single metric, making it particularly effective for evaluating unbalanced datasets. Defined in Equation (18), the F-measure is calculated as the harmonic mean of accuracy and precision. Its value ranges between 0 and 1, where 1 represents perfect agreement and 0 indicates no agreement.

$$F - \text{meature} = 2 \cdot \frac{\text{precision} \cdot \text{recall}}{\text{precision} + \text{recall}}$$

(18)

In the context of the above formula, the term Precision refers to accuracy, specifically the closeness of measurement values to one another, regardless of whether these values reflect the true reality or not.

**Precision measure**

The precision measure evaluates clustering accuracy based on the class labels by assessing the correct assignment of data points to their respective clusters. This measure accounts for both the items correctly assigned to their actual clusters and those correctly excluded from incorrect clusters. Mathematically, as defined in Equation (19), precision is calculated as the ratio of the sum of correctly clustered items and correctly excluded items to the total number of data points. This measure provides a clear indication of the clustering algorithm's effectiveness in accurately grouping the data.

$$Accuracy = \frac{TP + TN}{N} \quad (19)$$

**Normalized mutual information criterion**

In this method, clusters are evaluated using a sustainability criterion based on Normalized Mutual Information (NMI). The approach involves a mechanism to assess the stability of individual clusters independently of other clusters produced during the clustering process. This ensures that each cluster's stability is measured solely based on its consistency and reproducibility across multiple clustering iterations or sampling processes. To do so, suppose that we want to calculate the stability of cluster $C_i$. In this method, new data sets are formed by sampling and different clustering approaches are done on them. Then, it will be tried to answer this question: "Has this cluster also appeared in the clustering approaches or not?" To do so, a similarity measure between that cluster ($C_i$) and the initial clustering (P (D)) is suggested which is shown as $sim(C_i, P(D))$. Using this criterion, the similarity between that cluster and different clusterings done by sampling is calculated. The mean of the similarity criteria is then returned as the stability of the cluster $g_i(C_i, D)$. In fact, $sim(C_i, P(D))$ specifies the validity of cluster $C_i$ in P clustering on $D$ dataset. The $NMI$ relationship between P1 and P2 clustering is calculated with equation 20.

$$NMI = \frac{MI(P1,P2)}{\frac{-1}{2m}\left(\sum_{i=0}^{1} p_i \log \frac{p_i}{m} + \sum_{j=0}^{1} p_j \log \frac{p_j}{m}\right)} \quad (20)$$

In this equation, $p_{11}$ represents the number of common samples in $C^*$ and $C_i$. $p_{10}$ shows the number of common samples in $D/C^*$ and $C_i$. $p_{01}$ indicates the number of common samples in $C^*$ and $D/C_i$. $p_{00}$ indicates the number of common samples in $D/C^*$ and $D / C_i$. Moreover, m is the total number of samples. In fact, $p_i$. and $p_{i.}$ are respectively the total samples available in $C_i$ and $C^*$.

**ARI standard**

The consistency between U and C partitions can be shown by the probability matrix $M \in R^{K_c \times K_u}$ in which $K_c$ and $k$ are respectively the numbers of the clusters in partitions C and U. $M_{ij} = |C_i \cap U_j|$ is the number of available data in cluster I of partition $C$ and cluster j of partition $U$. For two partitions $C$ and $U$ the Rand Index includes the following information:

1. $a$: The number of pairs of data in similar clusters in $C$ and $U$.

2. $b$: The number of pairs of data in the similar clusters in $C$, but not in the similar cluster $U$.

3. $c$: The number of pairs of data in the similar clusters in $U$, but not in cluster $C$.

4. $d$: The number of pairs of data in different clusters in both partitions.

$$\frac{(a+b)}{(a+b+c+d)} \quad (21)$$

The criteria $a$ and $d$ can be considered as consistent standards while the criteria $b$ and $c$ can be considered as mismatches. Adjust Rand Index ($ARI$) is a criterion to measure the similarity between two partitions and is calculated based on the following formula:

$$ARI = \frac{\sum_{ij} C^2_{M_{ij}} - [\sum_i C^2_{M_{i-}} \cdot \sum_j C^2_{M-j}]/C^2_M}{\frac{1}{2}[\sum_i C^2_{M_{i-}} \cdot \sum_j C^2_{M-j}] - [\sum_i C^2_{M_{i-}} \cdot \sum_j C^2_{M-j}]/C^2_M} \quad (22)$$

In formula 22, $M_{ij}$ is the number of data items that exist in cluster i of partition C and cluster j of partition U. $M_i$ is the number of data items in cluster $i$ of partition $C$ ( the sum of the i$^{th}$ row of matrix M). $M - j$ is the number of data items in cluster $j$ of partition $U$ ( the sum of the j$^{th}$ column of matrix $M$).

**Experiments**

The Genetic MDL method was implemented and tested using Python 3.4. Its performance was compared against the following clustering algorithms: K-means, Single Linkage, Average Linkage, Complete Linkage, Ward Linkage, and FCM. The experiments were conducted over 100 independent runs of the program, and the results were reported as averages accompanied by their standard deviations. The algorithm rankings based on various validation criteria are presented in Tables 2 to 5 below.

**Evaluation of Genetic MDL**

In this section, the results obtained from 100 independent runs of each algorithm are presented. As shown in Table 2 and Figure 5, the algorithms were evaluated on thirteen datasets using the NMI criterion. The results demonstrate that, on average, across 100 independent implementations for each dataset, the Genetic MDL method consistently outperformed the other algorithms in terms of efficiency.

**Table 2.** Comparison of average normalized data of Genetic MDL method with the common clustering algorithms

| Data sets | Methods | | | | | | | | |
|---|---|---|---|---|---|---|---|---|---|
| | k-means | single | Average | Complete | Ward | Fcm | EPMDLGAO | EPAFGAO | ABMDLGAO |
| balance_scale | **12.39± 4.94** | 3.61± 0 | 6.6 ± 0 | 5.57± 0 | 8.55± 0 | 11.63± 8.48 | 9.28±7.77 | 8.26±1.44 | 9.69±9.140 |
| Halfring | 32.04 ± 0 | 6.65 ± 0 | 69.7 ± 0 | 50.18 ± 0 | 50.18 ± 0 | 32.04 ± 0 | 68.69±1.86 | 46.72±14.84 | **71.99±10.75** |
| Iris | 74.64 ± 4.32 | 76.12 ± 0 | 80.58 ± 0 | 72.24 ± 0 | 77.01 ± 0 | 74.96 ± 0 | 86.49±7.51 | 85.34±3.92 | **97.1±2.01** |
| Nbalance_scale | **11.51 ± 3.03** | 3.61 ± 0 | 6.24 ± 0 | 7.63 ± 0 | 2.97 ± 0 | 10.79 ± 9.32 | 8.47±.39 | 11.03±1.93 | 8.85±6 |
| Nbreast | 71.24 ± 0.51 | 1.82 ± 0 | 15.6 ± 0 | 70.28 ± 0 | 79.11 ± 0 | 71.16 ± 0 | 74.05±1.99 | 6.06±21.91 | **84.14±2.61** |
| Nbupa | 0.15 ± 0.01 | 1.36 ± 0 | 2.59 ± 0 | 0.07 ± 0 | 0.01 ± 0 | 0.81 ± 0 | 2.52±.82 | 1.73±1.4 | **6.29±15.7** |
| Ngalaxy | 28.61 ± 2.43 | 12.17 ± 0 | 31.61 ± 0 | 21.05 ± 0 | 28.29 ± 0 | 26.86 ± 0.02 | **39.55±4.38** | 24.96±2.94 | 28.06±5.57 |
| Nglass | 27.41 ± 1.99 | 11.96 ± 0 | 30.02 ± 0 | 18.4 ± 0 | 27.53 ± 0 | 30.83 ± 0.06 | 49.99±18.22 | 27.71±13.57 | **59.93±11.83** |
| Nionosphere | 12.5 ± 0 | 11.96 ± 0 | 2.59 ± 0 | 7.37 ± 0 | 13.28 ± 0 | **15.83 ± 1.19** | 14.65±4.57 | 10.62±4.45 | 12.44±1.27 |
| NSAHeart | 7.6 ± 0.38 | 5.19 ± 0 | 5.17 ± 0 | 1.85 ± 0 | 6.89 ± 0 | 8.6 ± 0 | **9.2±2.81** | 9.09±2.22 | 4.27±.67 |
| Nwine | 86.84 ± 5.26 | 5.83 ± 0 | 3.08 ± 0 | 61.44 ± 0 | 78.65 ± 0 | 48.39 ± 0 | **87.07±7.25** | 51.77±6.88 | 63.42±5.57 |
| NYeast | 29.86 ± 0.75 | 9.99 ± 0 | 12.04 ± 0 | 18.91 ± 0 | 28.73 ± 0 | 15.55 ± 0.92 | **31.67±5.65** | 29.22±13.26 | 26.57±40.53 |
| Wine | 42.13 ± 0.9 | 9.14 ± 0 | 41.58 ± 0 | 44.23 ± 0 | 41.61 ± 0 | 41.68 ± 0 | 77.17±4.31 | 51.24±3.75 | **89.1±2.75** |

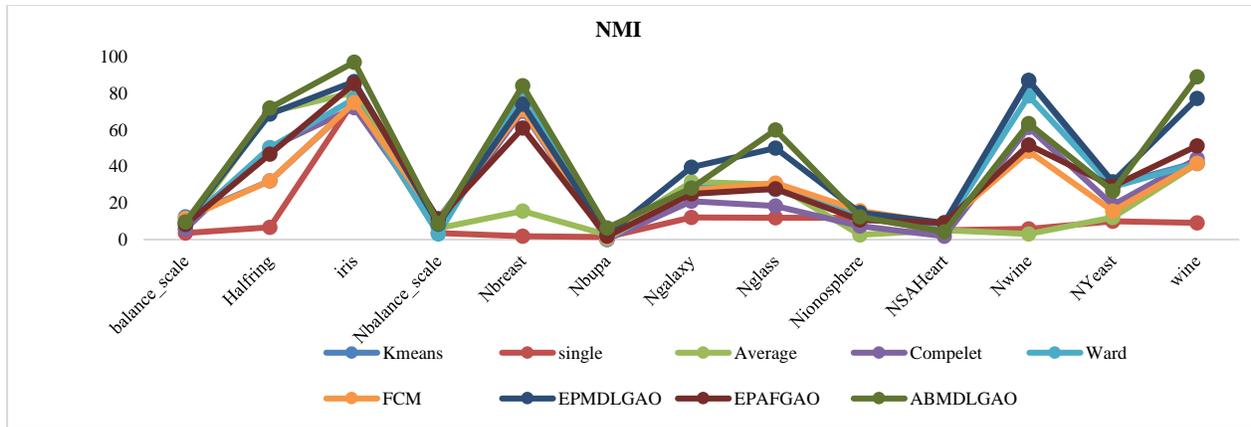

**Fig.5.** The NMI graph of clustering methods

As presented in Table 3 and Figure 6, the algorithms were assessed on thirteen datasets using the ARI criterion. The results, averaged over 100 independent runs for each dataset, indicate that the Genetic MDL method consistently achieved superior efficiency compared to the other algorithms.

**Table 3.** Comparison of average ARI of Genetic MDL method with common clustering algorithms

| Data sets | Methods | | | | | | | | |
|---|---|---|---|---|---|---|---|---|---|
| | K-means | Single | Average | Complete | Ward | Fcm | EPMDLGAO | EPAFGAO | ABMDLGAO |
| **balance_scale** | 13.24 ± 3.78 | 0.36± 0 | 9.2 ± 0 | 6.03 ± 0 | 11.98 ± 0 | 10.73 ± 8.31 | **13.5±.58** | 13.14±.71 | 12.65±.66 |
| **Halfring** | 23.87 ± 0 | 2.99 ± 0 | 78.07 ± 0 | 51.16 ± 0 | 51.16 ± 0 | 23.87 ± 0 | 94.31±4.3 | 58.94±3.18 | **96.29±5.39** |
| **Iris** | 63.85 ± 14.01 | 56.38 ± 0 | 75.92 ± 0 | 64.23 ± 0 | 73.12 ± 0 | 72.94 ± 0 | 78.5±4.82 | **96.24±4.25** | 73.44±4.21 |
| **Nbalance_scale** | 13.55 ± 3.88 | 0.36 ± 0 | 8.54 ± 0 | 8.3 ± 0 | 3.55 ± 0 | 14.82 ± 10.99 | **15.01±.64** | 8.42±.52 | 14.62±.53 |
| **Nbreast** | 81.61 ± 0.44 | 0.25 ± 0 | 9.63 ± 0 | 79.75 ± 0 | 86.35 ± 0 | 81.38 ± 0 | **91.7±3.28** | 85.45±1.09 | 78.56±.71 |
| **Nbupa** | -.069 ± 0.02 | -0.16 ± 0 | -0.46 ± 0 | -0.44 ± 0 | -0.3 ± 0 | -0.6 ± 0 | 1.62±.3 | 1.57±.28 | **1.62±.54** |
| **Ngalaxy** | 11.17 ± 1.04 | -0.03 ± 0 | 13.4 ± 0 | 6.93 ± 0 | 10.59 ± 0 | 10.54 ± 0 | 4.16±.96 | **27.51±2.76** | 15.55±2.15 |
| **Nglass** | 19.04 ± 5.39 | 1.49 ± 0 | 1.95 ± 0 | 3.5 ± 0 | 13.51 ± 0 | 15.44 ± 0 | 33.63±3.65 | 33.45±4.33 | **34.76±2.78** |
| **Nionosphere** | 16.79 ± 0 | 0.45 ± 0 | 0.45 ± 0 | 11.45 ± 0 | 17.75 ± 0 | 15.87 ± 0 | 1.53±3.23 | **19.57±.63** | -1.23±.47 |
| **NSAHeart** | 6.57 ± 0.24 | -0.2 ± 0 | 1.55 ± 0 | 2.54 ± 0 | 7.98 ± 0 | 8.72 ± 0 | 3.15±2.43 | **10.13±5.13** | 9.31±1.82 |
| **Nwine** | 89.75 ± 0 | -0.68 ± 0 | -0.54 ± 0 | 57.71 ± 0 | 78.99 ± 0 | 89.75 ± 0 | 67.95±7.2 | **90.87±6.03** | 51.27±11.34 |
| **NYeast** | 17.47 ± 1.17 | 1 ± 0 | 1.87 ± 0 | 10.83 ± 0 | 16.92 ± 0 | 12.19 ± 0.73 | 44.81±3.03 | 45.1±3.49 | **48.49±4.63** |
| **Wine** | 35.75 ± 1.74 | 0.54 ± 0 | 29.26 ± 0 | 37.08 ± 0 | 36.84 ± 0 | 35.39 ± 0 | 46.32±4.59 | **53.26±8.35** | 41.66±3.33 |

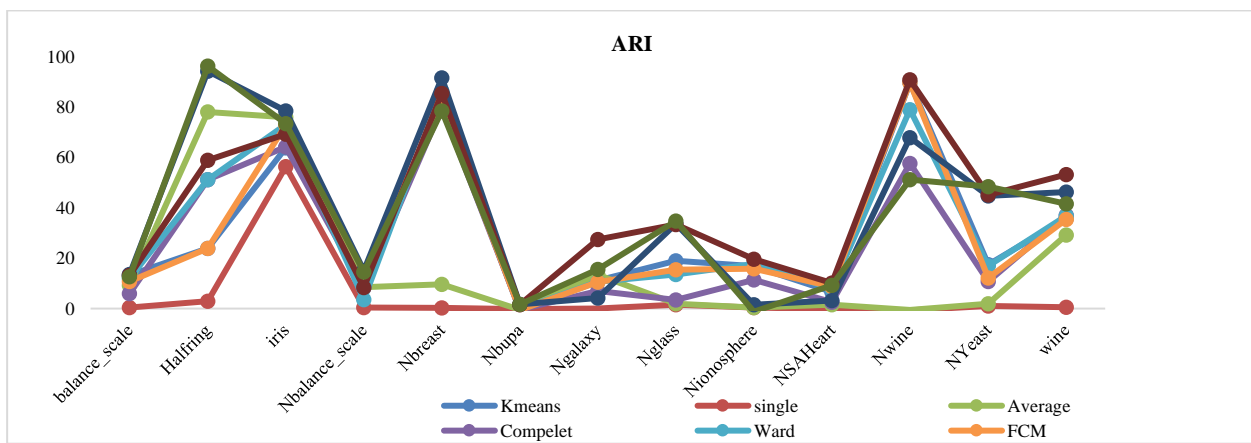

**Fig.6.** ARI graph of clustering methods

In Table 4 and Figure 7, the precision metric—which assesses the accuracy of the clustering method in assigning data points to their respective clusters—is presented. Based on the results averaged over 100 independent runs for each algorithm on the thirteen datasets, the Genetic MDL method demonstrated greater efficiency compared to the other methods.

Table 4. Comparison of the mean Accuracy of Genetic MDL with common clustering algorithms

| Data sets | Methods | | | | | | | | |
|---|---|---|---|---|---|---|---|---|---|
| | K-means | Single | Average | Complete | Ward | Fcm | EPMDLGAO | EPAFGAO | ABMDLGAO |
| balance_scale | 51.75 ± 2.71 | 3.61 ± 0 | 53.12 ± 0 | 50.24 ± 0 | 55.04 ± 0 | 53.65 ± 8.37 | 53.81±4.68 | 47.64±6.79 | **56.66±9.84** |
| Halfring | 74.5 ± 0 | 75.75 ± 0 | 94.75 ± 0 | 86 ± 0 | 86 ± 0 | 74.5 ± 0 | 93.64±2.39 | 80.20±11.63 | **95.44±3.95** |
| Iris | 86.58 ± 9.36 | 68 ± 0 | 90.67 ± 0 | 84 ± 0 | 89.33 ± 0 | 89.33 ± 0 | 87.51±9.21 | 82.23±5.63 | **91.17±6.5** |
| Nbalance_scale | 51.94 ±3.25 | 46.4 ± 0 | 52.16 ± 0 | 49.28 ± 0 | 43.2 ± 0 | 50.48 ± 7.14 | 48.85±4.97 | **53.79±5.22** | 47.49±6.33 |
| Nbreast | 95.25 ± 0.13 | 65.15 ± 0 | 70.13 ± 0 | 94.73 ± 0 | 96.49 ± 0 | 95.17 ± 0 | **96.7±2.61** | 89.17±9.44 | 93.89±3.71 |
| Nbupa | 54.58 ± 0.13 | 57.68 ± 0 | 57.1 ± 0 | 55.94 ± 0 | 55.65 ± 0 | 51.3 ± 0 | **64.84±6.69** | 56.5±7.84 | 58.33±9.15 |
| Ngalaxy | 30.88 ± 1.73 | 25.08 ± 0 | 33.44 ± 0 | 28.79 ± 0 | 29.1 ± 0 | 29.32 ± 0.14 | 31.67±3.99 | 28.76±1.73 | **36.34±4.11** |
| Nglass | 47.55 ± 4.45 | 36.45 ± 0 | 37.85 ± 0 | 40.65 ± 0 | 42.06 ± 0 | 40.53 ± 0.2 | 47.01±5.13 | **51.19±6.31** | 48.12±4.21 |
| Nionosphere | 70.66 ± 0 | 64.39 ± 0 | 64.39 ± 0 | 67.24 ± 0 | 71.23 ± 0 | 70.09 ± 0 | 66.65±8.7 | **71.73±7.28** | 68.82±6.63 |
| NSAHeart | 63.1 ± 0.24 | 65.15 ± 0 | 66.23 ± 0 | 63.1 ± 0 | 64.29 ± 0 | 64.93 ± 0.02 | 66.32±3.73 | **69.32±4.08** | 66.38±4.57 |
| Nwine | 96.63 ± 0 | 37.64 ± 0 | 38.76 ± 0 | 83.71 ± 0 | 92.7 ± 0 | 96.63 ± 0 | **97.13±.69** | 67.45±.8 | 88.79±1.49 |
| NYeast | 42.09 ± 2.37 | 31.74 ± 0 | 32.41 ± 0 | 35.92 ± 0 | 41.58 ± 0 | 33.09 ± 0 | 55.39±2.73 | 56.41±9.04 | **61.44±2.31** |
| Wine | 65 ± 6.54 | 42.7 ± 0 | 61.24 ± 0 | 67.42 ± 0 | 69.66 ± 0 | 68.54 ± 0 | 79.32±14.58 | 59.83±6.25 | **86.37±9.32** |

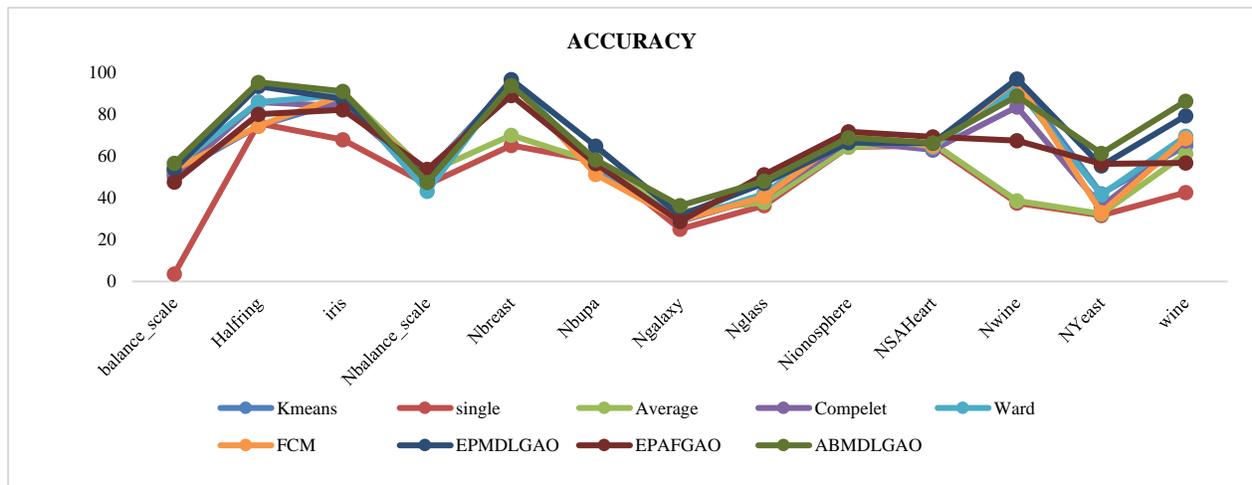

**Fig.7.** Accuracy graph of clustering methods

As shown in Table 5, the algorithms were evaluated on thirteen datasets using the Fisher standard. Based on the results averaged over 100 independent runs for each dataset, the Genetic MDL method consistently outperformed the other algorithms in terms of efficiency.

Table 5. Comparison of the mean Fisher standard method of MDL Genetic with common clustering algorithms

| Data sets | Methods | | | | | | | | |
|---|---|---|---|---|---|---|---|---|---|
| | K-means | Single | Average | Complete | Ward | Fcm | EPMDLGAO | EPAFGAO | ABMDLGAO |
| balance_scale | 55.9 ± 3.05 | 89.83 ± 0 | 53.16 ± 0 | 51.56 ± 0 | 53.4 ± 0 | 50.61± 6.25 | 56.99±10.43 | **88.09±7.02** | 54.96±9.77 |
| Halfring | 78.65 ± 0 | 92.83 ± 0 | 92.97 ± 0 | 86.14 ± 0 | 86.14 ± 0 | 78.37±0 | 84.53±12.52 | 84.44±6.36 | **95.19±6.34** |
| Iris | 88.22 ± 5.41 | 90.55 ± 0 | **91.68 ± 0** | 86.03 ± 0 | 90.29 ± 0 | 89.8±0 | 90.15±2.75 | 88.09±7.13 | 91.08±2.89 |
| Nbalance_scale | 54.94 ± 2.54 | 89.83 ± 0 | 52.83± 0 | 51.82± 0 | 47.3± 0 | 49.86±0 | 55.61±7.96 | **91.12±3.7** | 61.94±8.21 |
| Nbreast | 94.75 ± 0.14 | 90.36 ± 0 | 68.23± 0 | 94.22± 0 | 96.3± 0 | 94.67±0 | **97.49±2.58** | 84.23±8.36 | 95.74±4.75 |
| Nbupa | 55.38 ± 0.05 | 83.01 ± 0 | 61.74± 0 | 54.89± 0 | 54.2± 0 | 62.36±0 | **84.290±8.15** | 57.06±2.52 | 65.16±9.94 |
| Ngalaxy | 37.39 ± 2.73 | 66.53 ± 0 | 42.16± 0 | 35.32± 0 | 37.65± 0 | 31.43±0 | 45.66±1.47 | 49.81±11.98 | **68.55±8.29** |
| Nglass | 55.43 ± 3.41 | 82.8 ± 0 | 37.58± 0 | 41.76± 0 | 55.15± 0 | 50.42±0 | 70.36±1.61 | **84.76±6.23** | 69.65±1.05 |
| Nionosphere | 70.24 ± 0 | 90.03 ± 0 | 62.47± 0 | 65.65± 0 | 70.84± 0 | 69.79±0 | 72.62±1.25 | **90.19±4.54** | 77.93±6.87 |

**Algorithm ranking**

In this section, the ranking of each algorithm is presented based on the T-test results, derived from the average of 100 independent runs of each algorithm across 13 datasets for each criterion. Before discussing the rankings, a brief explanation of the T-test is provided.

**Introduction of the T test**

The T-test is a statistical method used to compare the performance of algorithms in machine learning. Assuming two algorithms, AAA and BBB, and a dataset ZZZ, the general process of this method is as follows:

1. Let P(A)P(A)P(A) and P(B)P(B)P(B) represent the precision values of algorithms AAA and BBB, respectively, on the dataset ZZZ.

2. Both algorithms are executed KKK times, and their results are compared with the real labels of the dataset.

3. A set of KKK differences is calculated by subtracting the precision values of the two algorithms for each test run, forming a set of differences {P(A)−P(B)}\{P(A) - P(B)\} {P(A)−P(B)}.

These differences are then analyzed statistically using the T-test to determine whether the performance of the two algorithms differs significantly. This comparison helps in ranking the algorithms based on their efficiency and accuracy.

$$P^{(1)} = P_A^{(1)} - P_B^{(1)} \ldots P^{(k)} = P_A^{(k)} - P_B^{(k)}$$

The samples are selected independently. With the null hypothesis ($H_0$: accuracy equivalent), t is calculated with the freedom degree k-1 as follows:

(23)
$$t = \frac{P^- \sqrt{K}}{\sqrt{\sum_{i=1}^{k} \frac{(p(i) - p^-)^2}{k-1}}}$$

where $P^- = \left(\frac{1}{k}\right) \sum_{i=1}^{k} P(i)$

In the final step of the T-test, the calculated $ttt$-value, approximated using the given formula, is compared with the critical $ttt$-value at a 0.05 significance level (5% error). If the calculated $ttt$-value exceeds the critical $ttt$-value, the difference between the algorithms is considered statistically significant; otherwise, it is deemed nonsignificant. The results of these comparisons, along with the algorithm rankings based on each criterion, are presented in Tables 6 to 9.

**Table 6.** Algorithm priorities based on ARI criterion

| Data sets | Methods | | | | | | | | | | | | |
|---|---|---|---|---|---|---|---|---|---|---|---|---|---|
| | balance_scale | Halfring | Iris | Nbalance_scale | Nbreast | Nbupa | Ngalaxy | Nglass | Nionosphere | NSAHeart | Nwine | NYeast | Wine |
| kmeans | 5 | -5 | -5 | 3 | 2 | -8 | 2 | 2 | 4 | 0 | 3 | 2 | -2 |
| Single | -8 | -8 | -8 | -8 | -8 | 2 | -6 | -8 | -6 | -8 | -8 | -8 | -8 |
| Average | -3 | 4 | 6 | 0 | -5 | -4 | 4 | -6 | -4 | -6 | -6 | -6 | -6 |
| Complete | -6 | -1 | -4 | -6 | -2 | -2 | -4 | -4 | 0 | -4 | -2 | -4 | -2 |
| Ward | -1 | -1 | 2 | -4 | 6 | 0 | -2 | -2 | 6 | 2 | 4 | 0 | 2 |
| Fcm | -2 | -5 | 1 | 5 | -1 | -6 | -2 | 0 | 2 | 4 | 7 | -2 | -2 |
| EPMDLGAO | 7 | 6 | 8 | 7 | 8 | 6 | -6 | 5 | -2 | -2 | 0 | 5 | 6 |
| EPAFGAO | 5 | 2 | -2 | -2 | 4 | 6 | 8 | 5 | 8 | 7 | 6 | 5 | 8 |
| ABMDLGAO | 3 | 8 | 2 | 5 | -4 | 6 | 6 | 8 | -8 | 7 | -4 | 8 | 4 |

**Table 7.** Algorithm priorities based on Fisher standard

| Data sets | Methods |
|---|---|

| | balance_scale | Halfring | Iris | Nbalance_scale | Nbreast | Nbupa | Ngalaxy | Nglass | Nionosphere | NSAHeart | Nwine | NYeast | Wine |
|---|---|---|---|---|---|---|---|---|---|---|---|---|---|
| Kmeans | 2 | -2 | -5 | 1 | 2 | -4 | -3 | -1 | -1 | 0 | 6 | 2 | -1 |
| Single | **8** | 6 | 6 | 7 | -2 | 7 | 6 | 6 | 7 | 6 | -6 | 2 | 2 |
| Average | -3 | 5 | **6** | -1 | -8 | 0 | 0 | -8 | -8 | -4 | -8 | -8 | -4 |
| Complete | -7 | -3 | -8 | -4 | -1 | -6 | -6 | -6 | -6 | -8 | -2 | -2 | -4 |
| Ward | -1 | -1 | 1 | -7 | 4 | -8 | -3 | -1 | -1 | -2 | 2 | 0 | 2 |
| Fcm | -7 | -8 | -1 | -6 | -2 | 2 | -8 | -4 | -4 | 2 | **7** | -6 | -3 |
| EPMDLGAO | 2 | -1 | 1 | -1 | **8** | 7 | 2 | 4 | 2 | -6 | 5 | -2 | 0 |
| EPAFGAO | 6 | -3 | -5 | 7 | -6 | -2 | 4 | **8** | 7 | 4 | -4 | **8** | 0 |
| ABMDLGAO | 0 | **7** | 5 | 4 | 5 | 4 | **8** | 2 | 4 | **8** | 0 | 6 | **8** |

**Table 8.** Algorithm priorities based on normalized mutual information criterion

| Data sets | Methods | | | | | | | | | | | | |
|---|---|---|---|---|---|---|---|---|---|---|---|---|---|
| | balance_scale | Halfring | Iris | Nbalance_scale | Nbreast | Nbupa | Ngalaxy | Nglass | Nionosphere | NSAHeart | Nwine | NYeast | Wine |
| kmeans | **7** | -5 | -5 | 6 | 0 | -4 | 2 | -2 | 1 | 2 | 7 | 5 | 0 |
| Single | -8 | -8 | -6 | -6 | -8 | 0 | -8 | -8 | -2 | -2 | -6 | -8 | -8 |
| Average | -4 | 6 | 2 | -4 | -6 | 5 | 6 | 1 | -8 | -4 | -8 | -6 | -6 |
| Complete | -6 | 0 | -6 | -2 | -3 | -6 | -6 | -6 | -6 | -8 | 0 | -2 | 0 |
| ward | 1 | 2 | 0 | -8 | 6 | -8 | 2 | -2 | 4 | 0 | 4 | 3 | -2 |
| Fcm | 6 | -5 | -3 | 5 | 0 | -2 | -2 | 4 | **8** | 4 | -4 | -4 | -2 |
| EPMDLGAO | 1 | 4 | 5 | 1 | 4 | 5 | **8** | 6 | 6 | **7** | **7** | 7 | 6 |
| EPAFGAO | 1 | -2 | 5 | 6 | -1 | 2 | -4 | -1 | -4 | 7 | -2 | 4 | 4 |
| ABMDLGAO | 2 | **8** | **8** | 2 | **8** | **8** | 2 | **8** | 1 | -6 | 2 | 1 | **8** |

**Table 9.** Algorithm priorities based on precision criterion

| Data sets | Methods | | | | | | | | | | | | |
|---|---|---|---|---|---|---|---|---|---|---|---|---|---|
| | balance_scale | Halfring | Iris | Nbalance_scale | Nbreast | Nbupa | Ngalaxy | Nglass | Nionosphere | NSAHeart | Nwine | NYeast | Wine |
| kmeans | -2 | 0 | -1 | 4 | 4 | -6 | 3 | 4 | **7** | -6 | 2 | 2 | -2 |
| Single | -8 | -6 | -8 | -5 | -8 | -2 | -8 | -8 | -8 | -2 | -8 | -8 | -8 |
| Average | 2 | 6 | 3 | 5 | -6 | 2 | 6 | -6 | -6 | 2 | -6 | -6 | -4 |
| Complete | -4 | -1 | -4 | 0 | 3 | 1 | -3 | 0 | -1 | -8 | -2 | -2 | 0 |
| ward | 6 | -6 | 5 | -8 | 3 | -1 | -5 | -2 | 3 | 0 | 4 | 0 | 2 |
| Fcm | 3 | -2 | 3 | 1 | 3 | -6 | 0 | -4 | 1 | -2 | 6 | -4 | 4 |
| EPMDLGAO | 2 | 4 | 1 | -1 | 7 | **8** | 3 | 4 | -2 | 4 | **8** | 5 | 6 |
| EPAFGAO | -6 | -2 | -6 | **8** | 4 | 1 | -4 | **8** | 6 | **8** | -4 | 5 | -6 |
| ABMDLGAO | **7** | **7** | **7** | -4 | -2 | 3 | **8** | 4 | 0 | 4 | 0 | **8** | **8** |

**SUMMARY AND CONCLUSION**

In this study, the AWDL approach, an extended version of the MDL principle, was applied to clustering datasets. The key advantage of this method lies in its ability to consider both the internal and external information of the data. By weighting the attributes of the datasets, the AWDL approach assigns special importance to attribute value coefficients during clustering, reducing biases toward input clusters and producing more stable clustering results. Additionally, the proposed method incorporates multiple evaluation functions for assessing candidate solutions, enhancing its adaptability to diverse datasets. Experimental results on 13 standard datasets, evaluated using four widely accepted validation metrics—Fisher, accuracy, NMI, and ARI—demonstrated the superior efficiency of the proposed method compared to conventional clustering algorithms such as Linkage Single, Linkage Average, Linkage Complete, Linkage Ward, K-means, and FCM.

The AWDL approach is particularly promising for applications in medical and health sciences, where clustering methods are often used to analyze complex and high-dimensional data. For example, in EEG (electroencephalogram) and ECG (electrocardiogram) signal analysis, clustering can help identify patterns associated with neurological disorders or cardiac abnormalities, enabling early diagnosis and personalized treatment [45-46]. Similarly, in spatial transcriptomics, clustering is critical for grouping cells based on their gene expression profiles and spatial locations, aiding in the study of tissue architecture and disease progression, such as in cancer research. The stability and bias-resilient properties of AWDL make it particularly well-suited for these applications, where accurate clustering directly impacts clinical insights and outcomes.

Furthermore, the AWDL method holds significant potential for use in deep learning, where clustering can play a vital role in tasks such as feature extraction, unsupervised pretraining, and data augmentation [47-48]. Clustering methods like AWDL can be integrated into autoencoders, variational autoencoders (VAEs), or contrastive learning frameworks to identify latent structures in the data, improving representation learning. Additionally, the ability to handle diverse datasets and produce stable clusters makes AWDL an excellent choice for applications in domains like computer vision, natural language processing, and bioinformatics, where clustering is often used to preprocess data or interpret deep learning model outputs. By enhancing clustering performance, AWDL can contribute to developing more robust and efficient deep learning pipelines.

**Future Work**

One potential avenue for advancing the assessment of clustering methods is the development of a comprehensive dataset encompassing diverse data types with varying characteristics to serve as a benchmark for testing clustering algorithms. Additionally, building upon the present research, it is recommended to explore the use of other heuristic and meta-heuristic optimization techniques, such as Ant Colony Optimization, Bee Algorithms, and Frog Leap Algorithm, for refining the initial solution, alongside the genetic optimization algorithm employed in this study.